\documentclass[10pt,twocolumn,letterpaper]{article}

\usepackage{cvpr}
\usepackage{times}
\usepackage{epsfig}
\usepackage{graphicx}
\usepackage{amsmath}
\usepackage{amssymb}
\usepackage{booktabs}

\usepackage[pagebackref=true,breaklinks=true,letterpaper=true,colorlinks,bookmarks=false,urlcolor=red,citecolor=cyan]{hyperref}
\usepackage{caption}
\usepackage{enumitem}
\setitemize{noitemsep,topsep=0pt,parsep=0pt,partopsep=0pt}
\usepackage{authblk}

\cvprfinalcopy 

\def\cvprPaperID{66666} 

\begin{document}

\title{Cloth Interactive Transformer for Virtual Try-On}


\author{Bin Ren$^{1,2}$,
        Hao Tang$^3$, 
	Fanyang Meng$^4$,
	Runwei Ding$^4$,
 	Philip H.S. Torr$^5$,
	Nicu Sebe$^{1}$ \\
$^1$University of Trento, 
$^2$University of Pisa,
$^3$ETH Zürich,
$^4$Peng Cheng Laboratory, \\
$^5$University of Oxford
}

\maketitle

\begin{abstract}
The 2D image-based virtual try-on has aroused increased interest from the multimedia and computer vision fields due to its enormous commercial value. Nevertheless, most existing image-based virtual try-on approaches directly combine the person-identity representation and the in-shop clothing items without taking their mutual correlations into consideration. Moreover, these methods are commonly established on pure convolutional neural networks (CNNs) architectures which are not simple to capture the long-range correlations among the input pixels. As a result, it generally results in inconsistent results. To alleviate these issues, in this paper, we propose a novel two-stage cloth interactive transformer (CIT) method for the virtual try-on task. During the first stage, we design a CIT matching block, aiming to precisely capture the long-range correlations between the cloth-agnostic person information and the in-shop cloth information. Consequently, it makes the warped in-shop clothing items look more natural in appearance. In the second stage, we put forth a CIT reasoning block for establishing global mutual interactive dependencies among person representation, the warped clothing item, and the corresponding warped cloth mask. The empirical results, based on mutual dependencies, demonstrate that the final try-on results are more realistic. Substantial empirical results on a public fashion dataset illustrate that the suggested CIT attains competitive virtual try-on performance.
\end{abstract}
\section{Introduction}
\label{sec:intro}
Virtual try-on (VTON), derived from fashion editing~\cite{zhu2017your,men2020controllable,yu2019vtnfp,choi2021viton}, aims to transfer a desired in-shop clothing item onto a customer's body. If properly resolved, VTON will provide a time and energy-saving shopping experience in our everyday life. In practice, VTON has already been deployed in some big-brand clothing stores or e-commerce shopping applications owing to its convenience~\cite{morelli2022dress,lee2022high,he2022style,fincato2021viton}.
 
However, most of the existing methods are designed based on 3D models pipelines~\cite{alldieck2018video,lahner2018deepwrinkles, gundogdu2019garnet,guan2012drape,sekine2014virtual} and follow the conventions of traditional computer graphics. Despite the detailed results, these methods require considerable labor resources, a significant amount of time investment, and complex data acquisition such as multi-view videos or 3D scans~\cite{pons2017clothcap} that impede their widespread application. Alternatively, conditional generative adversarial Networks (GANs) based methods such as image-to-image translation or other image generation approaches~\cite{choi2018stargan,dong2019fw}, recently made some positive progress. However, there remain some obvious artifacts in the generated results. To enhance the results of 2D image-based VTON methods, the classic two-stage pipeline VTON~\cite{han2018viton} was proposed, utilizing the first stage to warp the in-shop clothing item to a desired deformation style, and in the second stage, the warped cloth is aligned to the body shape of a given customer. While the visual result looks better than previous methods, there is still a significant gap between the overall visual quality and plausible generation. Many approaches following this pipeline, \emph{i.e.}, CP-VTON~\cite{wang2018toward}, ACGPN~\cite{yang2020towards}, and CP-VTON+~\cite{minar2020cp} were proposed with improved performance. However, these methods are limited to plain texture or simple-style clothes. Their performance suffers when dealing with complicated cases like rich textures or complex patterns. To address this issue, Xu \textit{et al.} \cite{xu2021virtual} introduced an intermediate operation that takes the transformation of the target person's image into consideration. But the improvement in visual performance is achieved at the cost of a more complex network architecture, which is time-consuming for training such a model. In addition, we also noticed that most of the previous methods rarely pay enough attention to the correlations between two crucial input information, \emph{i.e.}, the cloth-agnostic person information and the in-shop cloth information. Hence, there will be some inevitable mismatch phenomena occurring in the warped in-shop cloths, and consequently, it degrades the quality of the final try-on results. Moreover, for VTON, it's essential for a model to learn where to sample the pixels in the cloth image and where to reallocate them in the human body region. Hence, modeling the long-range dependencies is essential for achieving realistic try-on results. However, most of the previous methods use pure convolutional neural networks (CNNs) for modeling long-range dependencies. Since the pure CNN-based methods struggle to establish the long-range dependencies due to the design nature of the convolutional kernels, the final try-on performance is decreased.

\begin{figure}[!t] \small
	\centering
	\includegraphics[width=1\linewidth]{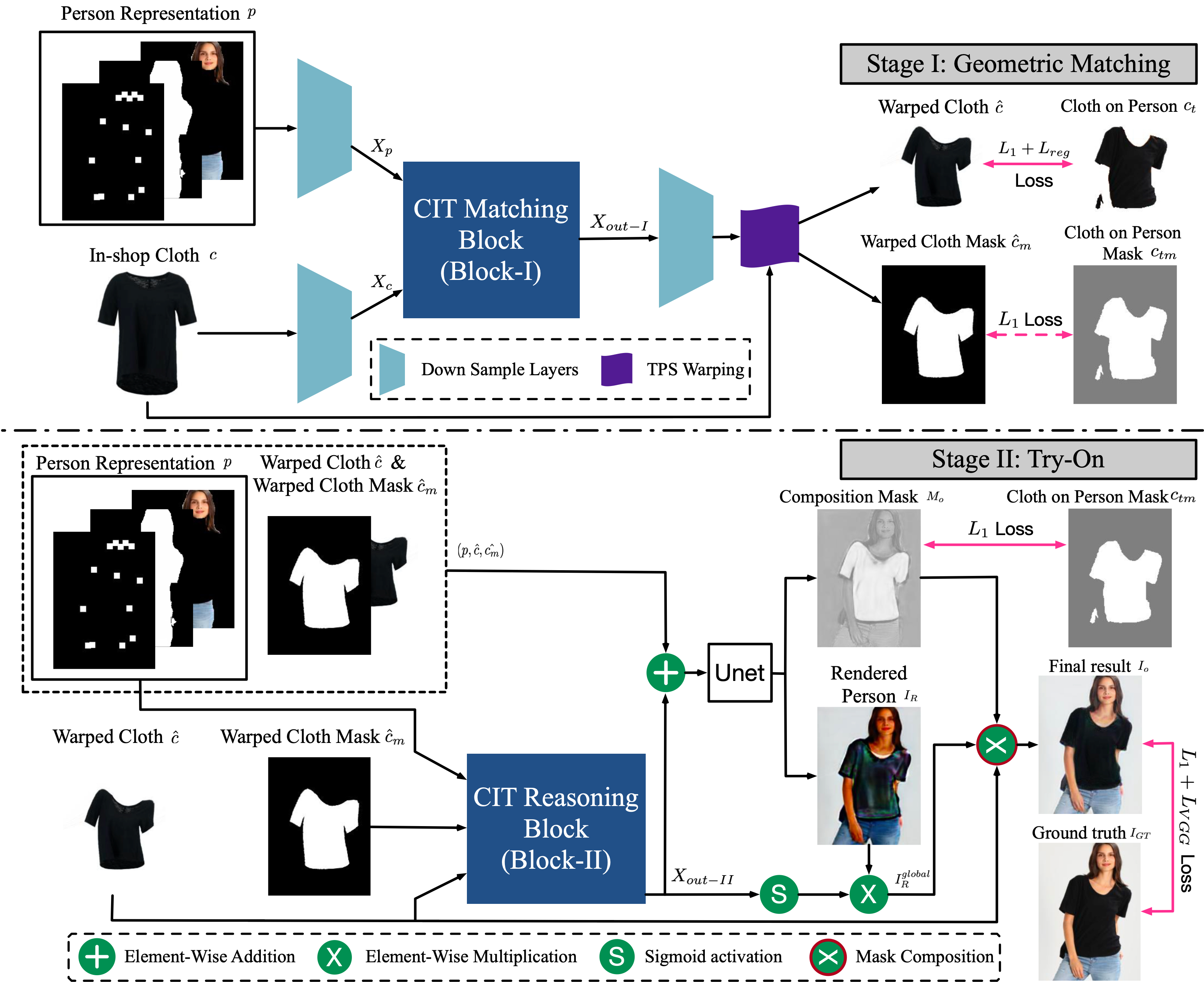}
	\caption{The overall architecture of the proposed CIT for virtual try-on. The upper part is the Geometric Matching stage for warping the in-shop clothing items, while the bottom part is the Try-On stage for synthesizing the final try-on image of the person.}
 	\vspace{-0.2cm}
	\label{fig:framework}
\end{figure}

Based on these observations, we assume that it would be advantageous to model the cloth-agnostic person information guided by the corresponding in-shop clothing information and vice versa. In addition, a better manner for capturing the long-range dependencies is also essential. To this end, based on the classic two-stage pipeline like VTON's~\cite{han2018viton}, in this paper, we propose a novel Cloth Interactive Transformer (CIT) method to address the aforementioned limitations. The overall architecture of the proposed CIT is depicted in Figure~\ref{fig:framework}.

In the first stage (\emph{i.e.}, geometric matching stage), we design a CIT matching block that models long-range relations between the person and clothing representations interactively. Concurrently, a valuable correlation map is generated to boost the performance of the thin-plate spline (TPS) transform~\cite{bookstein1989principal}. Unlike the traditional hand-crafted shape context matching strategies~\cite{mikolajczyk2002affine,lowe2004distinctive,schmid1997local}, which are only suitable for a certain feature type, the proposed CIT matching block ($Block$-I) has learnable features and can model long-range correlations via the cross-attention transformer encoders. As a result, the warped cloth becomes more natural and can fit a wearer's pose and shape more accurately. In the second stage (\emph{i.e.}, Try-On stage), unlike previous methods~\cite{wang2018toward,minar2020cp} that treat the warped in-shop clothing item and its corresponding mask as a single input, we propose a novel CIT reasoning block (($Block$-II)) that takes as input three distinct information, \emph{i.e.}, the cloth-agnostic person representations, the warped clothing item, and the mask of the warped clothing item. Through the CIT reasoning block, a more precise correlation among these three input data can be established, and this correlation is further utilized for strengthening the mask composition process. In addition, such a co-relation also serves as an attention map to activate the rendered person image, making the final results clearer and more realistic.
 
More specifically, in the CIT matching block, our primary objective is to improve the ability to model the person feature by encoding the target in-shop cloth feature. Since the in-shop clothing item is non-rigid, it is difficult to directly learn the matching relationship only from the clothing item. Hence, we resort to the correlation between the person and the target in-shop clothing item. With the help of this learned correlation, the person-related feature can be refined indirectly by the in-shop clothing features and vice versa. Such an analysis is also considered in the CIT reasoning block among the three input data. With the CIT matching and the CIT reasoning blocks, the correlation between both the person and in-shop cloth are updated synchronously and interact with each other.

In summary, our contributions are as follows:

\begin{itemize}
	\item We design a novel two-stage (\emph{i.e.}, the geometric matching stage and the try-on stage) Cloth Interactive Transformer (CIT) method for the challenging virtual try-on task. CIT can well model the long-range interactive relations between the cloth-agnostic person representations and the in-shop clothing items. 	
	
	\item We propose a new two-modality CIT matching block in the first geometric matching stage, making the in-shop clothing item can be better warped in the desired direction.
	
	\item We introduce a new three-modality CIT reasoning block in the second try-on stage. Based on this block, the more precise long-range correlations among three inputs (\emph{i.e.}, the cloth-agnostic person representation, the warped clothing item, and the corresponding mask of the warped cloth) can be well considered. As a result, CIT is able to obtain more realistic try-on image results.
\end{itemize}

\section{Related Work}
\label{sec:related}
\noindent \textbf{Virtual Try-On (VTON),} as one of the most popular tasks within the fashion area, has been widely studied by the research community due to its practical potential~\cite{fele2022c,cui2021dressing,bai2022single,ge2021disentangled,li2021toward,fincato2022transform}. Conventionally, this task was realized by computer graphic techniques, which build 3D models and render the output images via the precise control of geometric transformations or physical constraints~\cite{ehara2006texture,brouet2012design,chen2016synthesizing,guan2012drape,yuan2013mixed,sekhavat2016privacy}. By using 3D measurements or representations, these methods can generate promising results for VTON, but the additional requirements such as 3D scanning equipment, computation resources, and heavy labor are not negligible.

Compared to 3D-based methods, 2D generative adversarial networks (GANs) based methods are more applicable to online shopping scenarios. Jetchev and Bergmann \cite{jetchev2017conditional} proposed a conditional GAN to swap fashion articles with only 2D images. Another interesting GAN-based method SWAPGAN~\cite{liu2019swapgan} solved the VITON task in an end-to-end manner, but it utilizes 3 generators, and the balance among all generators is hard to control. Gu \emph{et al.}~\cite{gu2020toward} proposed a GAN-based image transformation strategy that can automatically learn the mapping from a combination of pose and text to a target fashion image. However, this method didn't consider pose variations, it also required the availability of the paired images of both the in-shop clothes and the wearer during inference. Hence their applicability in practical scenarios is limited. Unlike the previous 3D-based or GAN-based methods, VITON~\cite{han2018viton} tackled this problem with a coarse-to-fine architecture, which first computed a shape context~\cite{belongie2002shape} thin-plate spline (TPS) transformation~\cite{bookstein1989principal} for warping an in-shop clothing item on the target person and then blended the warped clothing item onto a given person. Note that the TPS transformation is a commonly used method of transforming a source image into a target image. It works by using a predefined set of constraints that define how the source image should be warped and transformed into the target image. The constraints are based on a set of surface points on the source image and their corresponding points on the target image. The warping algorithm then uses these points to calculate a set of transformations that will transform the source image into the target image. The transformations include scaling, rotation, translation, and shearing. The result is a warped version of the source image which accurately reflects the target image. This warping method used in the VTON task is to make the in-shop clothing item well warped toward the body shape of a given person. However, VITON~\cite{han2018viton} utilized hand-craft shape-context features for conducting the TPS transformation, which is not only time-consuming but also not robust when facing new samples. As an improvement, CP-VTON~\cite{wang2018toward} and CP-VTON+~\cite{minar2020cp} adopted the learnable TPS transformation method proposed by~\cite{rocco2017convolutional} via a convolutional geometric matcher. Although the correlation between the person and the in-shape clothing features was established by such a differentiable TPS transformation, and the generated try-on results are better, there are still obvious artifacts when facing heavy occlusions, rich texture, or large transformations. ACGPN~\cite{yang2020towards} was proposed to tackle these issues. Compared to CP-VTON, ACGPN used a semantic generation module to generate a semantic alignment of spatial layout. It also introduced the second-order difference constraint based on TPS. Though the performance is improved, the problem is still similar to previous methods~\cite{han2018viton,wang2018toward,minar2020cp} because they didn't consider the global long-range interactive correlations between the person representation and the in-shop clothing item. Recently, Chopra~\emph{et al.}~\cite{chopra2021zflow} proposed to solve the VTON task with a gated appearance flow. Though better results are achieved, the need of modeling the 3D geometric priors makes the overall procedure more complex.

To alleviate these problems, we propose a two-stage Cloth Interactive Transformer (CIT) method for the virtual try-on task. In particular, the proposed CIT can well capture the long-range dependencies in both stages. As a result, our method generates sharper and more realistic try-on images.

\noindent \textbf{Long-Range Dependence Modeling.}
Although CNN-based structures have shown excellent representation ability in various vision tasks such as classification, segmentation, and so on. The long-range dependencies are still hard to be established due to the limited receptive fields of the convolution kernels. For example, a convolutional kernel usually focuses on local neighbors (e.g., 3$\times$3 or 5$\times$5), while long-range relations would require the response at a position as a weighted sum of the features at all other positions. This limitation raises huge challenges for many applications where long-range relationships are needed.

To overcome this limitation, the attention mechanism~\cite{tang2020bipartite,tang2020xinggan,vaswani2017attention} has been widely used in vision tasks with CNN architectures though it was initially designed for natural language processing tasks. In addition, non-local neural networks~\cite{wang2018non} were designed based on the self-attention mechanism, allowing the model to capture the long-distance dependencies in the feature maps. However, this approach suffers from high memory and computation costs. \cite{schlemper2019attention} proposed an attention gate model to increase the sensitivity of a base model. Besides, the multilayer perceptrons (MLP) are also proposed for modeling the long-range relation, but it may heavily affect the efficiency~\cite{ren2021cascaded,tolstikhin2021mlp}. Moreover, Transformer~\cite{vaswani2017attention} was first introduced for neural machine translation tasks because it can model long-range dependencies in sequence-to-sequence tasks and captures the relations between arbitrary positions in the given sequence. Unlike previous CNN-based methods, Transformers are built solely on self-attention operations, which are strong in modeling the global context. After Transformers demonstrated their overwhelming power on a broad range of language tasks (\emph{e.g.}, text classification, machine translation, or question answering~\cite{khan2022transformers,devlin2018bert,ott2019fairseq,strubell2018linguistically,ren2023masked}). Recently, Transformer-based frameworks have also shown their effectiveness on various vision tasks~\cite{liu2022breaking}. In particular, vision Transformer (ViT)~\cite{dosovitskiy2020image,tsai2019MULT} splits the image into patches and models the correlation between these patches as sequences, and then the core self-attention module of ViT is stacked for modeling the long-range dependencies. 

To this end, in this paper, we also utilize a vision Transformer for handling the long-range dependencies among the cloth-agnostic person representation, the in-shop clothing item (for both the original cloth and the warped cloth), and the corresponding mask of the warped clothing item in a novel cross-modal manner.
\section{Cloth Interactive Transformer}
\label{sec:method}

\begin{figure*}[!t] \small
	\centering
	\includegraphics[width=1\linewidth]{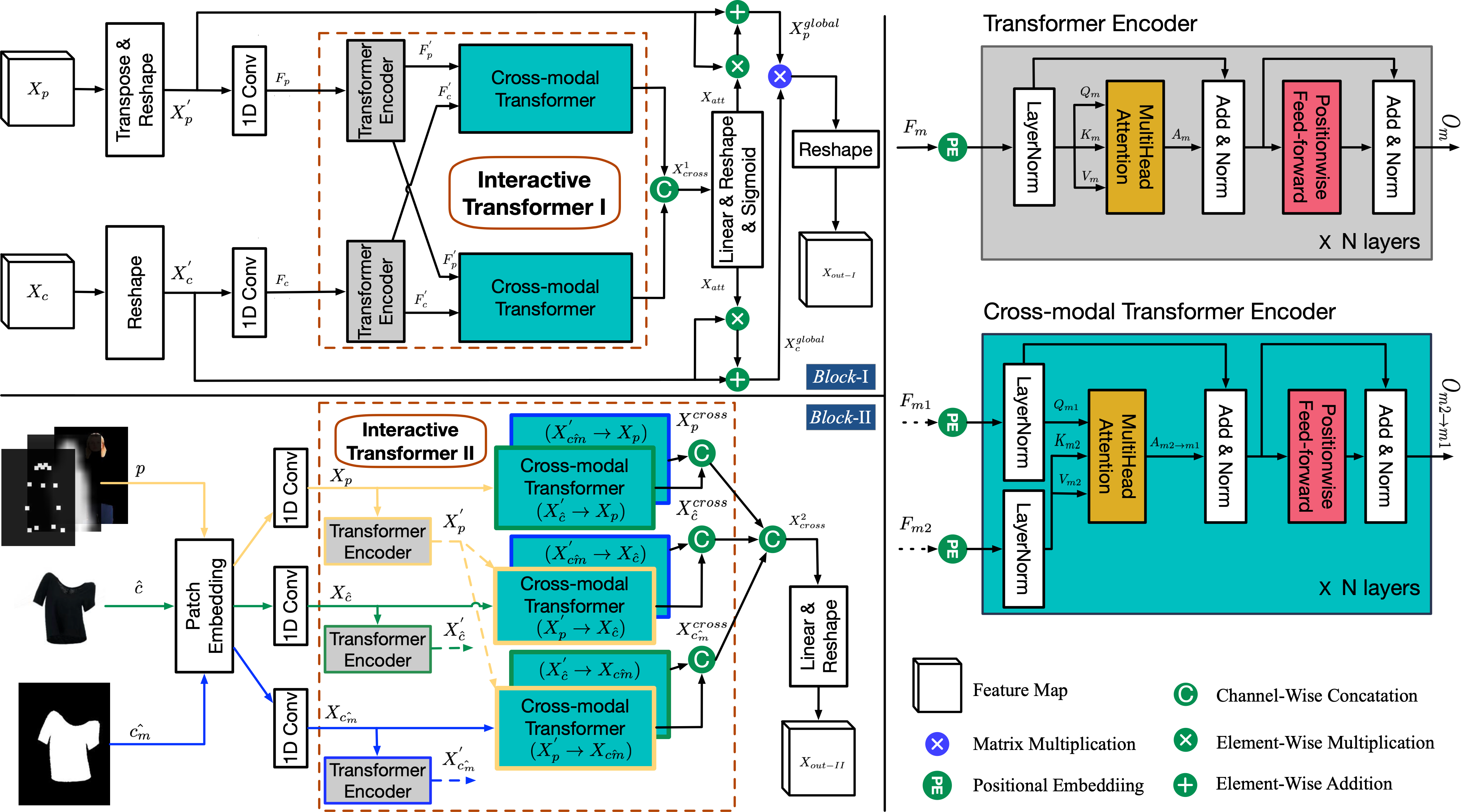}
	\caption{The key components of the proposed Cloth Interactive Transformer (CIT) for virtual try-on. The upper area on the left is the CIT Matching Block ($Block$-I), while the bottom area on the left indicates the CIT Reasoning Block ($Block$-II). On the right, the normal Transformer encoder and the proposed cross-modal Transformer encoder are shown in detail.}
	\label{fig:key-parts}
\end{figure*}

In this section, we first give an overall introduction and the necessary notations of the proposed Cloth Interactive Transformer (CIT) method for virtual try-on in~\ref{sublabel:overview}. Then we introduce the core modules (\emph{i.e.}, the Interactive Transformer I \& II) of CIT in Section~\ref{sublabel:interactive}. Based on the interactive transformer -I \& -II, we show the details of the CIT matching block (Block-I) and the CIT reasoning block (Block-II) in Section~\ref{sublabel:block1} and Section~\ref{sublabel:block2}, respectively. Finally, the optimization objectives of the proposed CIT for both stages are described in detail in Section~\ref{sublabel:loss}.

\subsection{Overview and Notations}
\label{sublabel:overview}
For the 2D image-based VTON task, the target in-shop clothing item is different from the source clothing item that is worn by a given person. Specifically, given a person image $I{\in}{\mathbb{R}^{3 \times h \times w}}$ and one in-shop clothing image $c{\in}{\mathbb{R}^{3 \times h \times w}}$. Our goal is to generate the image $I_{o} {\in}{\mathbb{R}^{3 \times h \times w}}$ where a person $I$ wears the cloth~$c$. Hence, what we need to do first is to reduce the side effects of the source clothes, like color, texture, or shape. Meanwhile, it's also necessary to preserve the information about the given person as much as possible, including the person's face, hair, body shape, and pose. To this end, we adopt the same pipeline as~\cite{han2018viton} for the person representation $p$ from $I$. It contains three components, the 18-channel feature maps for the human pose, the 1-channel feature map for the body shape, and the 3-channel RGB image. Note that the RGB image contains only the reserved regions of a person (\emph{i.e.}, face, hair, and lower parts of the person) for maintaining the identity of this person. 

The basic structure of the proposed CIT is in a two-stage (\emph{i.e.}, see the geometric matching stage and the try-on stage in Figure~\ref{fig:framework}) pipeline, which is also adopted by CP-VTON~\cite{wang2018toward} and CP-VTON+~\cite{minar2020cp}. In particular, the former takes as input the cloth-agnostic person representation $p$ and an in-shop clothing item $c$ to produce a warped cloth $\hat{c}$ and a warped mask $\hat{c_{m}}$ based on the given person's pose and shape. The latter uses the warped cloth $\hat{c}$, the corresponding warped mask $\hat{c_{m}}$ together with the person representation $p$ to generate the final person image with the worn in-shop cloth. In the first geometric matching stage, we propose a CIT matching block (Block-I, see the upper part in Figure~\ref{fig:key-parts} for the details), which takes the person feature $X_{p}$ and the in-shop cloth feature $X_{c}$ as inputs. Then $X_{p}$ and $X_{c}$ are generated by two similar feature extractors from $p$ and $c$, respectively (see the first geometric matching stage in Figure~\ref{fig:framework}). After that, it produces a correlation feature $X_{out-I}$ followed by a down-sample layer for regressing the parameter $\theta$. Note that $\theta$ is used for warping the original in-shop clothing $c$ to the target on-body style $\hat{c}$ via an interpolation method named the thin-plate spline (TPS) warping module~\cite{rocco2017convolutional}. Specifically, given two images with some corresponding control points in different positions, these control points can be well aligned from one (\emph{i.e.}, in the in-shop clothing item) to another (\emph{i.e.}, in the corresponding human body region) with the thin plate spline interpolation operation in a geometry estimation manner (\emph{i.e.}, local descriptor extraction, descriptor matching, transformation-related parameter estimation)~\cite{han2018viton}. 

In addition, the TPS operation we adopted in this paper is the same as the one used in CP-VTON~\cite{wang2018toward} from~\cite{rocco2017convolutional}. It first utilizes its differentiable modules to conduct a transformation by mimicking the geometry estimation procedure in a learnable manner from $c$ to the $\hat{c}$. Meanwhile, the corresponding mask $\hat{c_{m}}$ of $\hat{c}$ is also produced based on $\theta$ via TPS warping operation. In the second stage, we utilize the warped cloth $\hat{c}$ and the warped mask $\hat{c_{m}}$ together with the person representation $p$ as inputs of the CIT reasoning block (Block-II, see the bottom part in Figure~\ref{fig:key-parts} for the details). 
And the output $X_{out-II}$ of the CIT reasoning block is used to guide the final mask composition for generating more realistic try-on results.

\subsection{Interactive Transformer}
\label{sublabel:interactive}
Having leveraged the self-attention mechanism, Transformer is capable of modeling long-range dependencies. Given this inherent ability of the Transformer, we propose the innovative Interactive-Transformer for exploring the correlation between the person and the clothing item in the VTON task. There are two types of Interactive Transformers in the proposed CIT. The first version, \emph{i.e.}, Interactive Transformer I, is employed in the first geometric matching stage. The second version, \emph{i.e.}, Interactive Transformer II, is utilized in the second try-on stage. They are based on the basic Transformer encoders and the cross-modal Transformer encoders, and their detailed description is depicted in Figure~\ref{fig:key-parts}.

Regarding a standard Transformer encoder, a positional embedding is initially added to the input feature as elucidated in \cite{vaswani2017attention}. This is beneficial for keeping the initial spatial relation of the input. After the position embedding, the input feature will be projected into queries $Q_{m}$, keys $K_{m}$, and values $V_{m}$ by a linear layer. Subsequently, the output of the attention layer $A_{m}$ is derived as:
\begin{equation}
\begin{aligned}
A_{m} = {\rm softmax}(\frac{Q_{m} K^{T}_{m}}{\sqrt{d}}) V_{m},
\label{eq:loss_3}
\end{aligned}
\end{equation}
where $d$ is the dimension of the query, key.

The aforementioned self-attention mechanism is usually employed for only one type of input data. However, in the two-stage VTON task, to accurately capture a more precise match between the information of the person and the cloth information, there are several pairs of correlations we can't overlook. Notably, in the geometric matching stage, we need to consider the correlation between the cloth-agnostic person representation $p$ and the in-shop clothing item $c$ since such a correlation is indispensable for producing a reasonable warped cloth $\hat{c}$. In the second try-on stage, there are three types of inputs \emph{i.e.}, $p$, $\hat{c}$, as well as $\hat{c_{m}}$. To proficiently model the long-range connection of each two of them (\emph{i.e.}, $p$ and $\hat{c}$, $p$ and $\hat{c_{m}}$, as well as $\hat{c}$ and $\hat{c_{m}}$) is also a crucial issue since a well-captured correlation usually yields a good match between the person's body and the in-shop cloth.

Based on such an observation, instead of using only the self-attention layer in a Transformer encoder for processing a single-modal input, we propose a cross-modal Transformer encoder based on a cross-attention mechanism. Note that we treat each kind of input as a certain single-modal input since each of them provides a specific input. For example, $p$ is for person identity, $c$ and $\hat{c}$ correspond to the texture, and $\hat{c_{m}}$ is related to the shape information. And the cross-attention is computed as follows:
\begin{equation}
\begin{aligned}
A_{m2 \to m1} = {\rm softmax}(\frac{Q_{m1} K^{T}_{m2}}{\sqrt{d}}) V_{m2},
\label{eq:loss_4}
\end{aligned}
\end{equation}
where we adopt the first input (\emph{i.e.}, person representation $p$) as query $Q_{m1}$, and the second input (\emph{i.e.}, the in-shop clothing item $c$) as the keys $K_{m2}$ and values $V_{m2}$. Based on such a cross-interactive manner, each kind of input keeps updating its sequence via the external information from the multi-head cross-attention module. As a result, one modality will be transformed into a different set of key/value pairs to interact with another modality.

\noindent \textbf{Interactive Transformer I} is shown in the red dash box of the upper area of Figure~\ref{fig:key-parts}. It consists of two regular Transformer encoders (depicted in gray) and two cross-modal Transformer encoders (depicted in light blue) that are directly applied to feature maps. We use $selfTrans(\cdot)$ and $crossTrans(\cdot)$ to indicate the operators of these two kinds of Transformer encoders. Given two input features $F_{p}$ and $F_{c}$ with 
dimension $(C, B, S)$. Note that the dimension of $F_{p}$ and $F_{c}$ are reshaped from input features $X_{p}$ and $X_{c}$ with original dimension $(B, C, H, W)$. Here $B$, $C$, $H$, $W$ denote the batch size, the number of channels, the height, and the width of the input features $X_{p}$ and $X_{c}$, $S = H \times W$ denotes the spatial dimension. Then each of them will go through its corresponding $N$-layer regular Transformer encoder first. After that we get the processed features $F_{p}^{'}$ and $F_{c}^{'}$ as follows:
\begin{equation}
\begin{aligned}
F_{p}^{'} &= selfTrans(F_{p}),\\
F_{c}^{'} &= selfTrans(F_{c}).
\label{eq:loss_5}
\end{aligned}
\end{equation}

Then the cross-modal Transformer encoder is used for modeling the cross-modal long-range correlations between $F_{p}^{'}$ and $F_{c}^{'}$:
\begin{equation}
\begin{aligned}
X_{cross}^{1} = {\rm cat} \Big{(}crossTrans(F_{p}^{'}, F_{c}^{'}),\\
crossTrans(F_{c}^{'}, F_{p}^{'}) \Big{)},
\end{aligned}
\end{equation}
here $crossTrans(X_{p}^{'}, X_{c}^{'})$ indicates that we utilize $X_{c}^{'}$ as the keys and values while we use $X_{p}^{'}$ as the queries. On the other hand, $crossTrans(X_{c}^{'}, X_{p}^{'})$ indicates that the keys and values are coming from $X_{p}^{'}$ and queries come from $X_{c}^{'}$. After concatenating the outputs from the two cross-modal Transformer encoders, we get the output $X_{cross}^{1}$ of the Interactive Transformer I. It can strengthen the correlation matching ability.

\noindent \textbf{Interactive Transformer II} is shown in the red dash box at the bottom area of Figure~\ref{fig:key-parts}. Similar to the Interactive-Transformer I, the Interactive Transformer II is also constructed by combining regular Transformer encoders and cross-modal Transformer encoders. 

The Interactive Transformer II is designed mainly for exploring the correlations between every two inputs among three total inputs (\emph{i.e.}, $p$, $\hat{c}$, and $\hat{c_{m}}$). In particular, we adopt 3 regular Transformer encoders and 6 cross-modal Transformer encoders for constructing the Interactive Transformer II. Note that for better illustration, we depict $X_{p}$, $X_{\hat{c}}$, as well as $X_{\hat{c_m}}$ and their corresponding information flows in yellow, green, and blue, respectively. 

Within the Interactive Transformer II, there are three input features \emph{i.e.}, $X_p$, $X_{\hat{c}}$, and $X_{\hat{c_{m}}}$. Each of them works as the Query element within its own branch while working as the Key and Value elements in the other two branches. Specifically, we take the feature $X_p$ (depicted in yellow in Figure~\ref{fig:key-parts}) that comes from person representation as the detailed introduction. Once we get $X_p$ after the 1D convolutional layer that is out of the red dash box. There are two pathways for $X_p$ to pass through. The first one is to directly let it meet two cross-modal Transformer encoders (\emph{i.e.}, the green-border cross-modal Transformer encoder between $X_{\hat{c}}^{'}$ and $X_p$, as well as the blue-border cross-modal Transformer encoder between $X_{\hat{c_{m}}}^{'}$ and $X_p$). Another one is to let $X_p$ pass through a regular Transformer encoder for producing the updated feature $X_{p}^{'}$. Note that here $(X_{\hat{c}}^{'} {\to} X_{p})$ within the green-border cross-modal Transformer encoder means we utilize $X_{p}$ as Query and $X_{\hat{c}}^{'}$ as Key and Value, while $(X_{\hat{c_m}}^{'} {\to} X_{p})$ within the blue-border cross-modal Transformer encoder indicates we use $X_{p}$ as Query and $X_{\hat{c_m}}^{'}$ as Key and Value. $X_{\hat{c}}^{'}$ and $X_{\hat{c_m}}^{'}$ are the updated features from $X_{\hat{c}}$ and $X_{\hat{c_m}}$ after their corresponding regular Transformer encoders. We formulate such procedures of the first yellow branch as follows:
\begin{equation}
\begin{aligned}
X_{p}^{cross} = {\rm cat} \Big{(} crossTrans(X_{p}^{'},~X_{\hat{c}}^{'}),\\
crossTrans(X_{p}^{'}, X_{\hat{c_m}}^{'})\Big{)}
\label{eq:loss_7}
\end{aligned}
\end{equation}

Similarly, we also get the output of the middle green branch $X_{\hat{c}}^{cross}$ and the output of the bottom blue branch $X_{\hat{c_m}}^{cross}$. Finally, the overall output of the Interactive Transformer II is:
\begin{equation}
\begin{aligned}
X_{cross}^{2} = {\rm cat}(X_{p}^{cross}, X_{\hat{c}}^{cross}, X_{\hat{c_m}}^{cross}).
\label{eq:loss_8}
\end{aligned}
\end{equation}

\subsection{CIT Matching Block}
\label{sublabel:block1}
Based on our Interactive Transformer I, we propose the CIT Matching block (Block-I) to boost the performance of the TPS transformation by strengthening the long-range correlation between 
$X_{p}{\in} \mathbb{R}^{(B{\times}C{\times}H{\times}W)}$ and $X_{c}{\in} \mathbb{R}^{(B{\times}C{\times}H{\times}W)}$. Here $B$, $C$, $H$, and $W$ indicate batch size, channel number, and the height and width of a given feature. To utilize the Transformer encoder for modeling long-range dependencies, we first adjust the dimensions of $X_p$ and $X_c$ from $(B, C, H, W)$ to $(B, C, S)$, forming the $X_p^{'}$ and $X_c^{'}$. Note that $S=H \times W$. Besides, a 1D convolutional layer is also adopted to ensure that each element of each input sequence can get sufficient awareness of its neighborhood elements. When we get $F_p$ and $F_c$ after the convolutional layers, the proposed Interactive Transformer I is applied to $F_p$ and $F_c$ for capturing the long-range correlation between the person-related and the in-shop cloth-related features. As a result, we get the result (\emph{i.e.}, $X_{cross}^{1}$.) of the proposed CIT matching block. These procedures are depicted in Figure~\ref{fig:key-parts} with detailed annotations.

Instead of directly adding this long-range relation to features $X_{p}$ or $X_{c}$, we strengthen each of them by a global strengthened attention $X_{att}$ operation as follows:
\begin{equation}
\begin{aligned}
X_{(.)}^{global} = X_{(.)} + X_{(.)} \times X_{att},
\label{eq:loss_9}
\end{aligned}
\end{equation}
Here $\times$ means an element-wise multiplication, $(.)$ indicates that both features $X_{p}$ and $X_{c}$ follow the same form. Note that $X_{att}$ is produced from $X_{cross}^{1}$ by a linear projection and a sigmoid activation. Based on this operation, the element position relation of each input will be activated by the sigmoid activation function. In particular, when it is applied to the input feature as attention, both the position information of each element within each input and the correlation between two inputs can be kept in a balanced manner. Then a matrix multiplication between $X_{p}^{global}$ and $X_{c}^{global}$ is conducted. The output $X_{out-I}$ of the proposed CIT matching block is finally obtained after a reshape operation, which represents the improved correlation between the person and clothing features. These procedures can be defined as follows:
\begin{equation}
\begin{aligned}
x_{out-I} = {\rm Reshape}\Big{(} (X_{c}^{global})^{T} \times X_{p}^{global}\Big{)}.
\end{aligned}
\end{equation}

Here $X_{p}^{global}$ and $X_{c}^{global}$ have the same dimension $(B, C, S)$, the output $X_{out-I}$ is in dimension $(B, S, H, W)$.

\subsection{CIT Reasoning Block}
\label{sublabel:block2}
Previous methods, CP-VTON~\cite{wang2018toward} and CP-VTON+~\cite{minar2020cp}, first concatenate the person information $p$, the warped cloth information $\hat{c}$, and the warped clothing mask $\hat{c_m}$ together. Then the concatenated input is directly sent to one UNet model as a single input to generate a composition mask $M_{o}$ as well as a rendered person image $I_{R}$. However, such a rough concatenate operation may lead to coarse information matching, and consequently, it would be difficult to achieve a well-matched final try-on result. 

To this end, we propose the CIT Reasoning block ( Block-II) depicted in Figure~\ref{fig:key-parts}, aiming to model such more complicated correlations among $p$, $\hat{c}$, and $\hat{c_{m}}$. Firstly, we adopt the patch embedding operation~\cite{dosovitskiy2020image} to all these three inputs. Then each of them goes through a 1D convolutional layer to ensure the relation modeling of each element with its neighbor elements. After that, we get $X_p$, $X_{\hat{c}}$, and $X_{\hat{c_{m}}}$. To well capture the complicated long-range correlations among these features, we apply the proposed Interactive Transformer II to $X_p$, $X_{\hat{c}}$, as well as $X_{\hat{c_{m}}}$. Then the output $X_{out-II}$ of Interactive Transformer II is utilized to guide the final mask composition for a better generation as follows:
\begin{equation}
\begin{aligned}
I_{R}^{global} & = {\rm sigmoid}(X_{out-II}) \times I_{R},\\
I_{o} & = M_{o} \times \hat{c} + (1 - M_{o}) \times I_{R}^{global},
\label{eq:loss_11}
\end{aligned}
\end{equation}
here $sigmod$ indicates the Sigmod activation function.

\subsection{Optimization Objectives}
\label{sublabel:loss}
The first stage of CIT is trained with sample triplets $(p, c, c_m)$, while the second stage is trained with $(p, \hat{c}, \hat{c_m})$. In addition, in the first matching stage, we adopt the same optimization objectives as CP-VTON+~\cite{minar2020cp}:
\begin{equation}
\begin{aligned}
\mathcal{L}_{Matching} =  \mathcal{L}_{1}(\hat{c}, c_{t}) + \frac{1}{2}\mathcal{L}_{reg},
\label{eq:loss_1}
\end{aligned}
\end{equation}
where $\mathcal{L}_{1}$ indicates the pixel-wise L1 loss between the warped result $\hat{c}$ and the ground truth $c_{t}$. $\mathcal{L}_{reg}$ indicates the grid regularization loss, and it can be formalized as follows:

\begin{equation}
\begin{aligned}
\mathcal{L}_{reg}\left(G_{x}, G_{y}\right)=\sum_{i=-1,1} \sum_{x} \sum_{y}\left|G_{x}(x+i, y)-G_{x}(x, y)\right| \\ + 
\sum_{j=-1,1} \sum_{x} \sum_{y}\left|G_{y}(x, y+j)-G_{y}(x, y)\right|,
\end{aligned}
\label{eqn:all}
\end{equation} 
where $G_x$ and $G_y$ indicate the grid coordinates of the generated images along the x and the y directions.

In the second stage, the optimization objective is as follows:
\begin{equation}
\begin{aligned}
\mathcal{L}_{Try-on} = || I_{0} - I_{GT}||_{1} +  L_{VGG} +  ||M_{o} - c_{tm}||_{1}.
\label{eq:loss_2}
\end{aligned}
\end{equation}
The first item aims to minimize the discrepancy between the output $I_{o}$ and the ground truth $I_{GT}$. The second item, the VGG perceptual loss~\cite{johnson2016perceptual}, is a widely used loss item in image generation tasks. It is an alternative to pixel-wise losses and it attempts to be closer to the perceptual similarity of human beings. The VGG loss is based on the ReLU activation layers of the pre-trained 19-layer VGG network. It can be expressed as follows:
\begin{equation}
\begin{aligned}
\mathcal{L}_{VGG}=\frac{1}{W_{i, j} H_{i, j}} \sum_{x=1}^{W_{i, j}} \sum_{y=1}^{H_{i, j}}\Big{(}\phi_{i, j}\left(I_{GT}\right)_{x, y} - \\ \phi_{i, j}\left(I_{o}\right)_{x, y}\Big{)}^{2},
\end{aligned}
\end{equation}
where $W_{i,j}$ and $H_{i,j}$ describe the dimensions of the respective feature maps within the VGG network. $\phi_{i,j}$ indicates the feature map obtained by the j-th convolution before the i-th max pooling layer within the VGG19 network. The third item is used to encourage the composition mask $M_{o}$ to select the most suitable warped clothing mask $c_{tm}$ as much as possible.
\section{Experiments}
\label{sec:experiments}

\begin{figure*}[!t] \small
	\centering
	\includegraphics[width=1\linewidth]{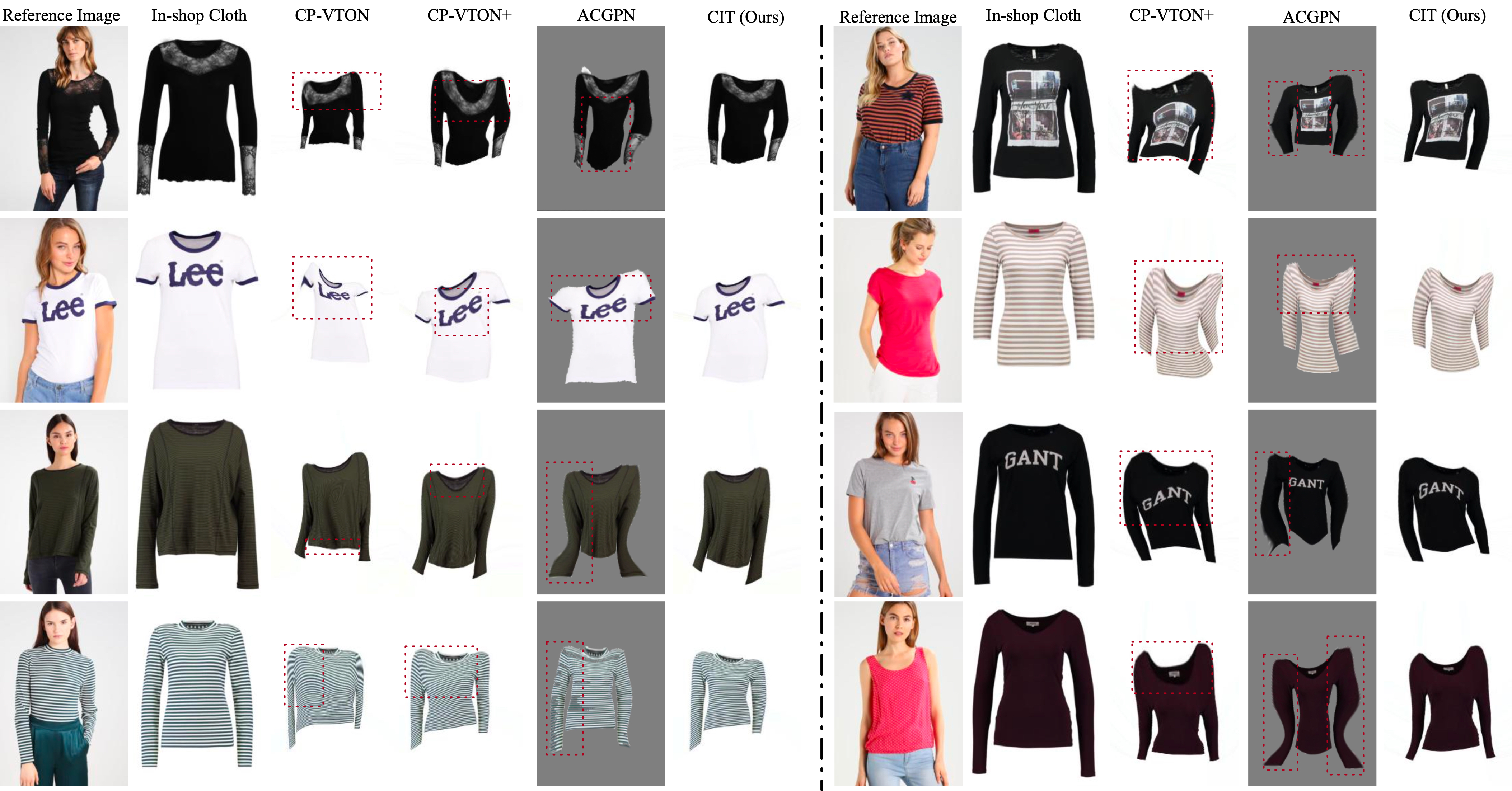} 
	\caption{Qualitative comparisons of the warped cloths by the proposed CIT-based geometric matching stage. The left is for the retry-on setting (\emph{i.e.}, in the same cloth) while the right is for the try-on setting (\emph{i.e.}, in different clothing items).}
	\vspace{-0.5cm}
	\label{fig:warp_result}
\end{figure*}

\noindent \textbf{Datasets.} 
We conduct all the experiments on the same dataset collected by Han \textit{et al.}~\cite{han2018viton} that used in VITON~\cite{han2018viton}, CP-VTON~\cite{wang2018toward}, and CP-VTON+~\cite{minar2020cp}. Note that due to copyright issues, we only use the reorganized version as the previous works~\cite{wang2018toward,minar2020cp} did. It contains around 19,000 front-view women and top clothing image pairs. Specifically, there are 16,253 cleaned pairs which are split into a training set and a validation set with 14,221 and 2,032 pairs, respectively. In the training set, the target cloth and the cloth worn by the wearer are the same. However, in the test stage, \textit{there are \textbf{two} kinds of test settings}. The first one is the same as the training settings, where the target clothing item and the clothing item worn by the wearer are the same (we refer to this case as a \textit{\textbf{retry-on setting}} because it is just like the wearer takes off the cloth first then retries this cloth on, hence we have the ground truth for this case). Another setting means that the target clothing item is different from the one worn by the wearer (we refer to this kind as the \textit{\textbf{try-on setting}}).

\noindent \textbf{Evaluation Metrics.}
To evaluate the performance of our method. We firstly adopt the Jaccard Score~\cite{jaccard1912distribution} for the \textit{\textbf{retry-on}} case (\emph{i.e.}, with ground truth) in the first stage. We also follow~\cite{minar2020cp,yang2020towards,issenhuth2020not} to use the Structural Similarity (SSIM) \cite{wang2004image}, Learned Perceptual Image Patch Similarity (LPIPS)~\cite{zhang2018unreasonable}, Peak Signal-to-Noise Ratio (PSNR), Frechet Inception Distance (FID), and Kernel Inception Distance (KID) metrics in the second stage.
Note that we adopt the original human image with the original clothing item as the reference image for SSIM and LPIPS (the lower, the better), and the parsed segmentation area for the current upper clothing is used as the reference for calculating the JS score. For \textit{\textbf{try-on}} case (no ground truth), we evaluate the performance of our method and other state-of-the-art methods by the Inception Score (IS)~\cite{salimans2016improved}.

\noindent \textbf{Implementation Details.} 
For the geometric matching stage, we build 2 feature extractors (see the downsample layers shown in Figure~\ref{fig:framework}) with 4 2-strided downsampling convolutional layers followed by 2 1-strided ones (filter numbers: 64, 128, 256, 512, 512) to generate $X_p$ and $X_c$. Those two extractors are only different in input channels. The regression networks before the TPS warping operation contain 2 2-strided convolutional layers, 2 1-strided ones (filter numbers: 512, 256, 128, 64), and 1 FC layer with output size 50. For the try-on stage, the U-net used here consists of 6 2-strided down-sampling convolutional layers (filter numbers: 64, 128, 256, 512, 512, 512) and 6 up-sampling  (filter numbers: 512, 512, 256, 128, 64, 4). Each convolutional layer is followed by one InstanceNorm layer, and a Leaky Relu with slope is set to 0.2. Note that we stack 3 CIT encoders in both matching and reasoning blocks.

Our training settings are similar to CP-VTON and CP-VTON+. Both stages are trained for 200K steps with batch size 4. Moreover, for Adam optimizer, $\beta_{1}$ and $\beta_{2}$ are set to 0.5 and 0.999, respectively. The learning rate was firstly fixed at 0.0001 for the first 100K steps and then linearly decayed to zero for the rest of the steps. All input images are resized to $256 {\times} 192$, and the output images have the same resolution. The source code and trained models are available at \href{https://github.com/Amazingren/CIT}{https://github.com/Amazingren/CIT}.

\subsection{Qualitative Comparisons}
To validate the performance of the proposed CIT for virtual try-on, we first present the visualization results of both stages, including the warped clothing items and the final try-on person images. 

\begin{figure*}[!t] \small
	\centering
	\includegraphics[width=1\linewidth]{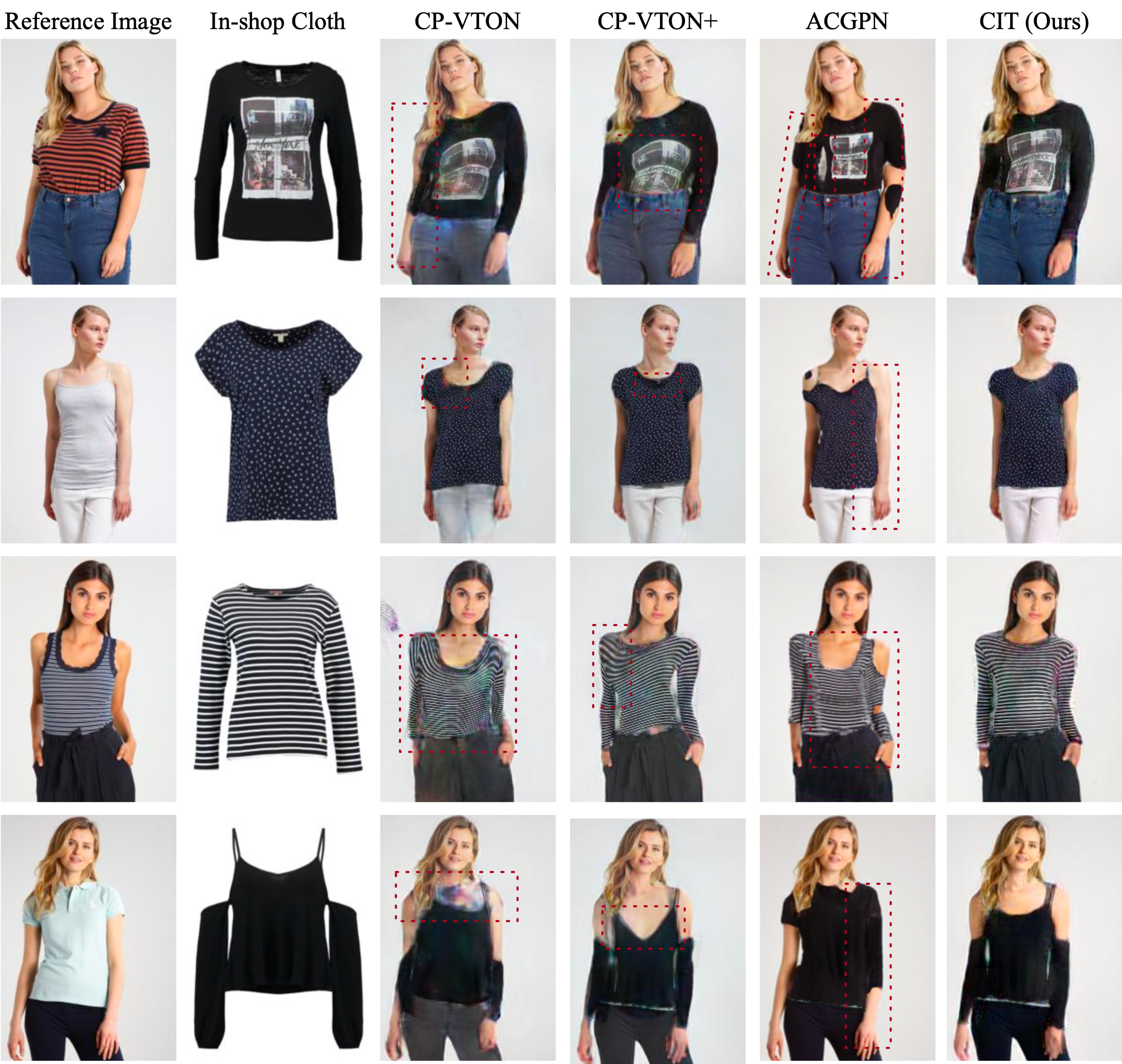} 
	\caption{Qualitative comparisons of different state-of-the-art methods.}
	\vspace{-0.5cm}
	\label{fig:sota_result}
\end{figure*}

\begin{figure*}[!t] \small
	\centering
	\includegraphics[width=1\linewidth]{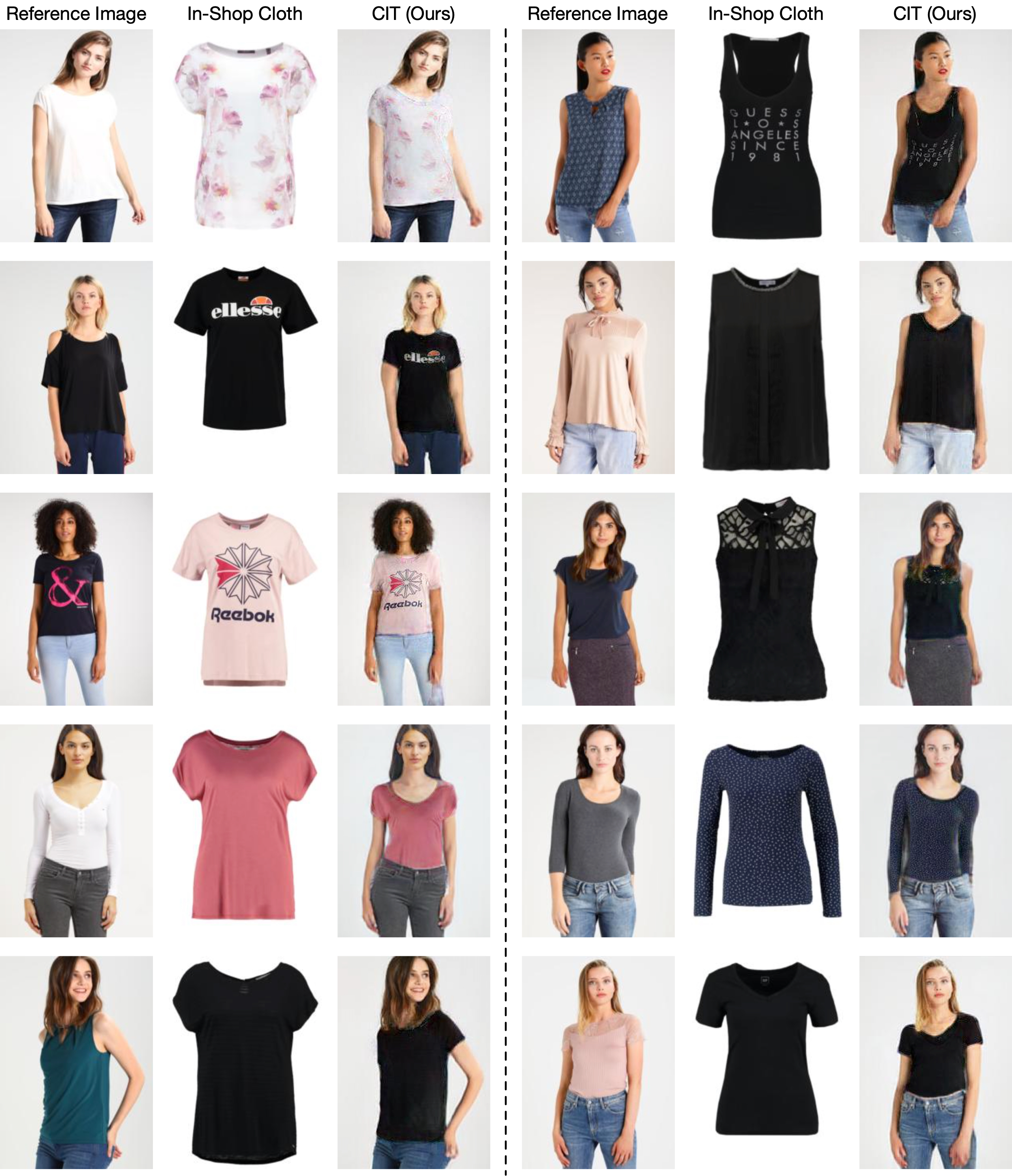} 
	\caption{More qualitative results of the proposed method.}
	\vspace{-0.5cm}
	\label{fig:more_vis}
\end{figure*}

\noindent \textbf{Comparison of Warping Results.}
We visualize the warped results of clothes for both retry-on and try-on settings in Figure~\ref{fig:warp_result}. Note that in this paper, our visualization method is similar to CP-VTON~\cite{wang2018toward}, and CP-VTON+~\cite{minar2020cp}. For ACGPN~\cite{yang2020towards}, we directly adopt its officially released code, and it produced the gray-background results. Figure~\ref{fig:warp_result} shows that the proposed CIT generates sharper and more realistic warped clothing items than the other methods like CP-VTON, CP-VTON+, and ACGPN. This typically happens for texture-rich cases, such as the case with line stripes (see the last row of the same-pair case and the second row of the different-pair case), or in the presence of logos (see the 2nd row of the same-pair case and the third row of the different-pair case), and so on. We marked out these obvious artifacts from other methods with red dash boxes in Figure~\ref{fig:warp_result}

\begin{table}[!t] \small
	\centering
	\caption{User study comparison on two questions. Q1 denotes `Which image is the most photo-realistic?', and Q2 denotes `Which image preserves the details of the in-shop clothing item the most?' in the user study }
		\begin{tabular}{lcc} \toprule
			Method & Q1 & Q2     \\ \midrule	
			CP-VTON~\cite{wang2018toward} &19.5 & 14.6 \\ 		
			CP-VTON+~\cite{minar2020cp}  &24.8 & 25.2 \\		
			ACGPN~\cite{yang2020towards}  &23.6 & 24.8\\
			CIT (Ours) & \textbf{32.1}  & \textbf{35.4}\\ 
			\bottomrule	
	\end{tabular}
	\label{tab:User_study}
\end{table}

\begin{table*}[!t] \small
	\centering
	\caption{Quantitative comparison in terms of JS, SSIM, LPIPS, PSNR, IS, FID, and KID evaluation metrics. For JS, SSIM, PSNR, and IS, the higher, the better, while for LPIPS, FID, and KID, the lower, the better. What's more, IS, FID, and KID are used to evaluate the unpaired \textit{\textbf{try-on}} setting while the rest are all for paired \textit{\textbf{retry-on}} setting. Note that the optimum results are \textbf{bolded} while the second-best results are \underline{underlined}}
		\begin{tabular}{lccccccc} \toprule
			Method &JS $\uparrow$ &SSIM$\uparrow$  & LPIPS$\downarrow$ & PSNR$\uparrow$ &  IS$\uparrow$ & FID$\downarrow$ & KID$\downarrow$  \\ \midrule	
			CP-VTON~\cite{wang2018toward}  &0.759 & 0.800 & 0.126 & 14.544 & 2.832 & 35.16 & 2.245\\ 
			CP-VTON+~\cite{minar2020cp}  &\textbf{0.812} & 0.817 & \underline{0.117} & 21.789 & \textbf{3.074} & 25.19 & 1.586 \\
            ACGPN~\cite{yang2020towards} & - & \textbf{0.846} & 0.121 & \underline{23.080} & 2.924 & \textbf{13.79} & \underline{0.818} \\
			CIT (Ours) & \underline{0.800} & \underline{0.827} & \textbf{0.115} & \textbf{23.464} & \underline{3.060} & \underline{13.97} & \textbf{0.761} \\ 
			\bottomrule	
	\end{tabular}
	\label{tab:JS_result}
\end{table*}

\noindent \textbf{Comparison of Try-On Results.}
Figure~\ref{fig:sota_result} shows the try-on results. We can see that the proposed CIT outperforms other methods. Specifically, the proposed CIT can keep the original clothing texture and its pattern as much as possible, and the final results are more realistic and natural. Compared to our method, other approaches display many artifacts, for example, the irregular logo pattern (the 1st row), the over-warped cloth texture (the 3rd row), and the ridiculous results for unique or complicated-style cloth (the last row). We also marked out these artifacts in red dash boxes in Figure~\ref{fig:sota_result}. In addition, more qualitative try-on results of the proposed method can be found in Figure~\ref{fig:more_vis}.

\noindent \textbf{User Study.}
We also evaluate the proposed CIT and other methods via a user study. We randomly select 120 sets of reference and target clothing images from the test dataset. Given the reference images and the target in-shop clothing items, 30 users are asked to choose the best outputs of our model and baselines (i.e., CP-VTON, CP-VTON+, and ACGPN) according to the two questions: (Q1) Which image is the most photo-realistic? (Q2) Which image preserves better the details of the target clothing? 
As shown in Table \ref{tab:User_study}, we can see that the proposed CIT achieves significantly better results than the other methods, which further demonstrates that our model generates more realistic images. And  CIT also preserves the details of the clothing items compared to the other methods.

\subsection{Quantitative Evaluation}
To further evaluate the performance of our CIT, we adopt five evaluation metrics, \emph{i.e.}, JS, SSIM, LPIPS, PSNR, IS, FID, and KID for numerical comparison. JS is to evaluate the quality of the warped mask in the first geometric matching stage with same-pair test samples, which is equivalent to the IoU metric used in CP-VTON+ but is more convenient for implementation. Note that we take cloth masks of the person as a reference image. Other metrics are designed to evaluate the performance of the second try-on stage.

The results of JS are shown in Table~\ref{tab:JS_result}. Though our CIT doesn't achieve the best JS score, our visual results presented in Figure~\ref{fig:warp_result} are the most reasonable ones. We think with the help of the proposed Interactive Transformer I in the CIT matching block, our method can learn more reasonable texture transformation patterns. This learned strong texture-focused transformation pattern might affect the shape alignment. This may come from the reason that the JS score only focuses on the shape alignment aspect between the ground truth mask and the warped clothing mask. Consequently, it can not well reveal the overall quality of the final generated human images. For instance, though Table~\ref{tab:JS_result} shows that CP-VTON+ has the best JS score of 0.812, which is higher than ours (0.800), the qualitative results show that our method is superior to CP-VTON+. So the shape-only related evaluation metrics \emph{i.e.}, JS or IoU, are not always indicating a better overall visual result. We also conducted ablation experiments in the following to support this conclusion (see the comparison results in Table \ref{tab:ablation} between B3 and B4). In addition, in terms of the realism of the generated images, the proposed CIT can also achieve the best KID and the second-best results, which confirm the effectiveness of our method.

\begin{figure*}[!t] \small
	\centering
	\includegraphics[width=1\linewidth]{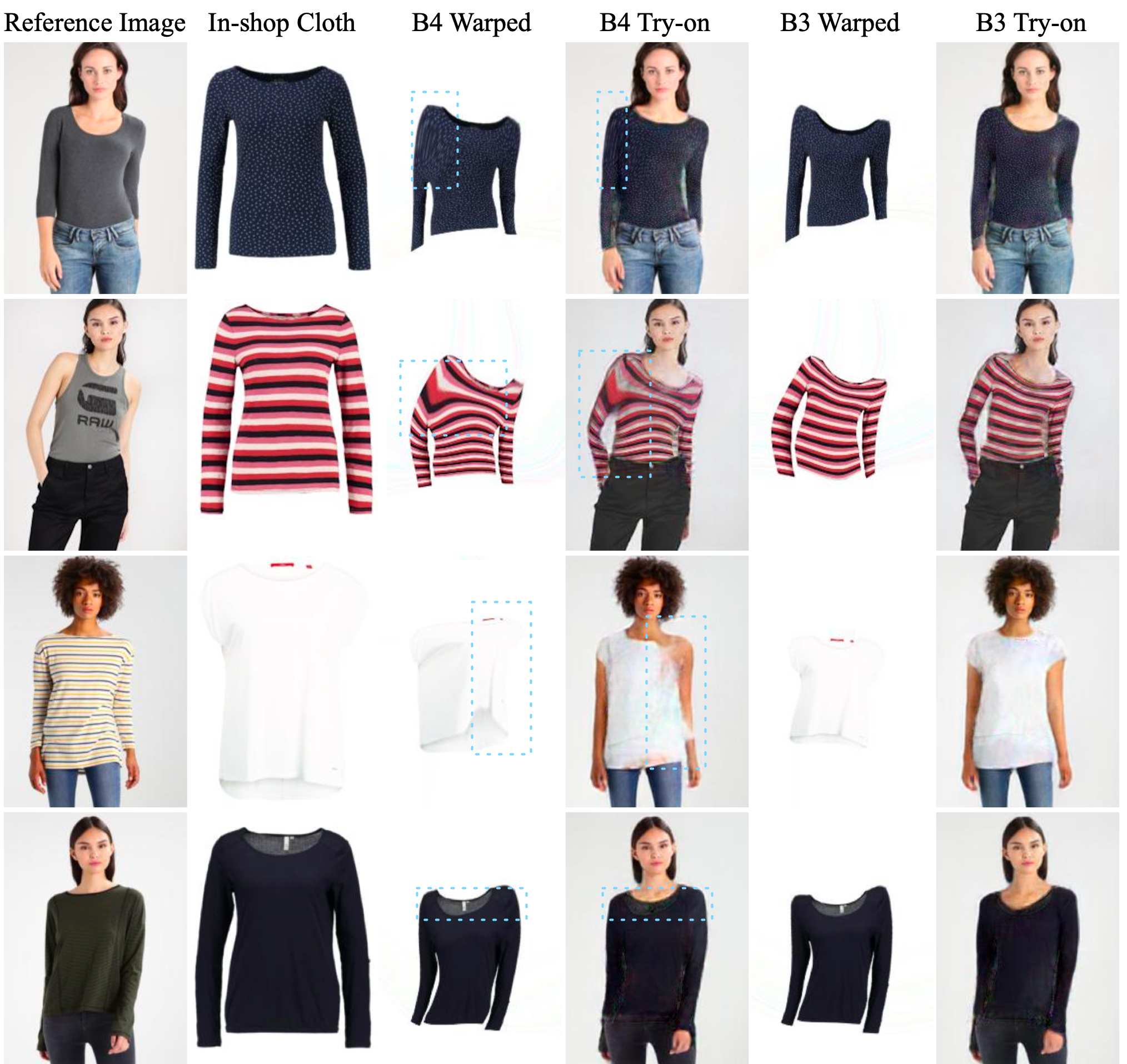} 
	\caption{Qualitative comparisons of ablation studies between B3 and B4.}
	\vspace{-0.2cm}
	\label{fig:b3b4}
\end{figure*}

\begin{table*}[!t] \small
	\centering
	\caption{Ablation studies of the proposed CIT for virtual try-on.}
		\begin{tabular}{lccccccccc} \toprule
			Method &JS$\uparrow$ & SSIM$\uparrow$ & LPIPS$\downarrow$ & IS$\uparrow$ & FID$\downarrow$ & KID$\downarrow$  & Q1 & Q2 \\ \midrule
			B0~\cite{minar2020cp}  &0.812 &0.817 & 0.117 & 3.074 & 25.19 & 1.586 & - & -\\ 
			B1 (CIT Matching only) &0.800 & 0.808 & 0.123 &  3.020 & 14.76 & 0.779 & - & -\\
			B2 (CIT Reasoning only)  & 0.812 & 0.821 & 0.125 & \textbf{3.105} & 14.87 & 0.784 & - & -\\
			B3 (Full: B1+B2) & 0.800 & 0.827 & 0.115 & 3.060 & \textbf{13.97} & \textbf{0.761} & \textbf{70.8} & \textbf{69.5 }\\
			B4 (Full + $L_{1}$ mask loss) & \textbf{0.813} & \textbf{0.829} & \textbf{0.110} & 3.005 & 14.32 & 0.788 & 29.2 & 30.5\\
			\bottomrule	
	\end{tabular}
	\label{tab:ablation}
\end{table*}

For the \textit{\textbf{retry-on}} setting, we adopt SSIM, PSNR, and LPIPS to evaluate the performance. The numerical results are shown in Table~\ref{tab:JS_result}. It can be seen that the proposed CIT achieves the best numerical evaluation results on SSIM and LPIPS compared to others.
For the \textit{\textbf{try-on}} setting, we use IS for evaluation. The results in Table~\ref{tab:JS_result} show that our CIT achieves
just a slightly lower IS score of 3.060 compared to 3.074 of CP-VTON+. We think the most possible reason for this phenomenon is that IS is an objective metric that usually was used to measure the quality of the generated images at the feature level based on image diversity and clarity. Hence, it may ignore some pixel-level properties.

Overall, though we do not obtain the best quantitative scores on JS and IS metrics, our proposed CIT can generate sharper and more realistic try-on images compared to others. For ACGPN, we also test the performance based on the corresponding official checkpoints. However, because the test set of ACGPN is different from others, so we only present its visual results in Figure~\ref{fig:warp_result} and Figure~\ref{fig:sota_result}.

\begin{figure*}[!t] \small
	\centering
	\includegraphics[width=1\linewidth]{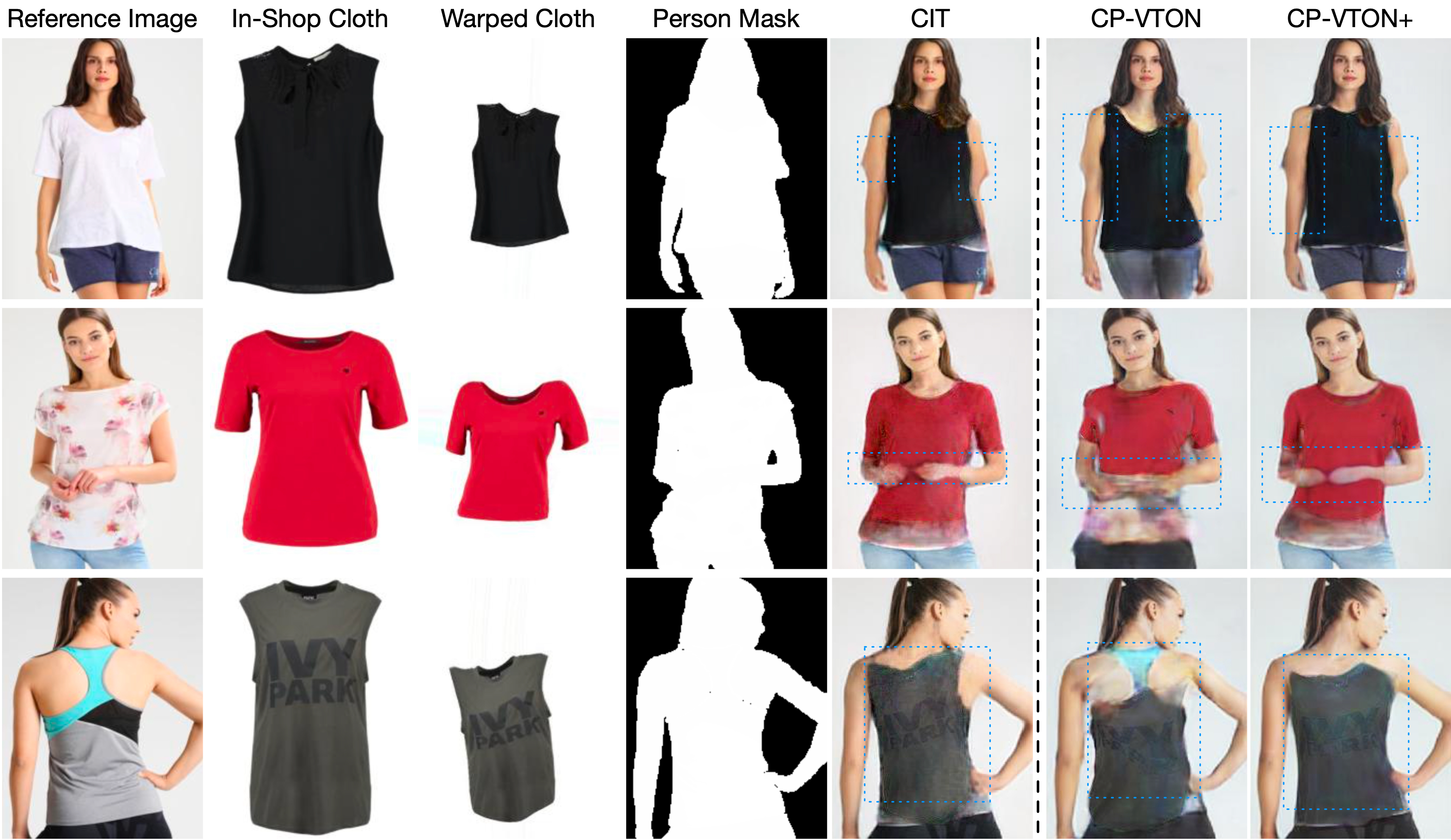} 
	\caption{Several failure cases of the proposed CIT, CP-VTON, and CP-VTON+.}
	\label{fig:failureCase}
\end{figure*}

\subsection{Ablation Study and Discussion}
To validate the effectiveness of each part of the proposed CIT, we conduct four ablation experiments (\emph{i.e.}, B1, B2, B3, and B4 in Table~\ref{tab:ablation}). CP-VTON+~\cite{minar2020cp} is adopted as the baseline (B0) of this paper. B1 means that we only use the proposed CIT matching block in the first geometric matching stage but keep the second stage the same as B0; B2 means that we only use the proposed CIT reasoning block in the second try-on stage while keeping the first stage the same as B0; B3 is the final version adopted in this paper, which contains both the proposed CIT matching and reasoning blocks; B4 is built based on B3, but with an extra $L_{1}$ loss item for providing more stricter constraints between the generated warped clothing mask $\hat{c_{m}}$ and the ground truth cloth mask $c_{tm}$ of the given person (depicted with red dash lines in Figure~\ref{fig:framework}). The setting of this experiment is designed to support the conclusion that a higher JS (or IoU) score doesn't indicate more enjoyable visual results since it ignores the texture or pattern aspect of the overall quality. The overall matching loss of B4 can be summarized as follows:
\begin{equation}
\begin{aligned}
L_{Matching} = L_{1}(\hat{c}, c_{t}) + L_{1}(\hat{c_{m}}, c_{tm}) + \frac{1}{2}\cdot L_{reg},
\label{eq:loss_12}
\end{aligned}
\end{equation}

From the comparison between B0~\cite{minar2020cp} and B1 in Table~\ref{tab:ablation}, though CP-VTON+ achieves a better JS score, the qualitative results presented in Figure~\ref{fig:warp_result} indicate that B1 can generate more reasonable and natural warped clothes (for both the retry-on and try-on settings). In other words, the proposed CIT Matching block can well capture more texture-related latent patterns with the help of the proposed Interactive Transformer I. The comparison between B0 and B2 shows that the numerical results, \emph{i.e.}, SSIM, and IS, are improved when we apply the proposed CIT reasoning block to the warped results from B0. This demonstrates that the proposed CIT reasoning block is more effective in generating the try-on results. B3, the combination of B1 and B2, is the final version of the proposed CIT. It produces not only the more natural warped clothing items but also achieves the more realistic try-on result. Hence, we think that for the two-stage 2D image-based VTON task, the JS or IoU metric focuses on only one aspect (\emph{i.e.}, shape) of the overall quality. Hence, the final try-on results are not always better when JS or IoU scores are higher. To support the above conclusion, we design B4 as a supplementary experiment based on B3. In Table~\ref{tab:ablation}, B4 obtains nearly all the best numerical results except the IS score. However, the visual comparison in Figure~\ref{fig:b3b4} shows that its virtual try-on results are far from satisfactory compared to B3. In addition, we asked 15 users for the user study according to the same questions asked in Table~\ref{tab:User_study} with randomly selected 50 image sets: (Q1) Which image is the most photo-realistic? (Q2) Which image preserves better the details of the target clothing? And the results shown in Table~\ref{tab:ablation} shows that B3 can generate more realistic images and preserves more details of the clothing items compared to B4. We also marked out the obvious artifacts region in Figure~\ref{fig:b3b4}.

\subsection{Failure Cases and Analysis}
Though impressive person try-on images can be generated by our CIT, there are still three kinds of inevitable common failure cases. We visualize them in Figure~\ref{fig:failureCase} with a comparison to both CP-VTON and CP-VTON+.

The first case (See the 1st row of Figure~\ref{fig:failureCase}) is that the difference between the clothing item in the reference image and the in-shop cloth is too large. Consequently, the mask of the person cannot well match the target in-shop clothing item. The second failure case comes from the self-occlusion problem, which leads to blurry ambiguity-prone generated images (See the 2nd row of Figure~\ref{fig:failureCase}). The third failure case (See the 3rd row of Figure~\ref{fig:failureCase}) derives from the drastic difference between the pose of the person and the sides of the in-shop cloth. This also leads to ambiguous results. In the first two cases, the main reason may be that the input data lack information in terms of whether the region of a human body should be covered with cloth or not. We propose to give further organization to the input data to remedy this issue, such as using more accurate segmentation maps or adopting more fine-grained human annotations.
For the last case, the 2D image-based method cannot completely capture such a complicated relationship between a person and a clothing item. We think that taking the 3D input data, such as body mesh, and 3D clothing items, into consideration may alleviate such a problem. 
\section{Conclusion}
\label{sec:conclusion}
In this paper, we propose a novel two-stage Cloth Interactive Transformer (CIT) method for the 2D image-based virtual try-on task. In the first stage, we introduce an interactive Transformer matching block, which is able to accurately model the global long-range correlations when warping a cloth through the thin-plate spline transformation. Consequently, the quality of the warped clothing item can be more realistic in terms of texture. We also present a transformer-based reasoning block in the second stage for modeling the mutual interactive relations, which can be utilized to further improve the rendering process, resulting in more realistic try-on results. Extensive experiments in terms of quantitative and qualitative comparisons validate that the proposed CIT achieves new competitive performance.

\noindent \textbf{Acknowledgments.} This work was supported by the National Ph.D. in Artificial Intelligence for Society Program of Italy, the MUR PNRR
project FAIR (PE00000013) funded by the NextGenerationEU and the EU H2020 AI4Media Project under Grant 951911.

{\small
\bibliographystyle{ieee_fullname}
\bibliography{egbib}

\begin{thebibliography}{10}\itemsep=-1pt

\bibitem{alldieck2018video}
Thiemo Alldieck, Marcus Magnor, Weipeng Xu, Christian Theobalt, and Gerard
  Pons-Moll.
\newblock Video based reconstruction of 3d people models.
\newblock In {\em Conference on Computer Vision and Pattern Recognition}, pages
  8387--8397, 2018.

\bibitem{bai2022single}
Shuai Bai, Huiling Zhou, Zhikang Li, Chang Zhou, and Hongxia Yang.
\newblock Single stage virtual try-on via deformable attention flows.
\newblock In {\em European Conference on Computer Vision}, pages 409--425.
  Springer, 2022.

\bibitem{belongie2002shape}
Serge Belongie, Jitendra Malik, and Jan Puzicha.
\newblock Shape matching and object recognition using shape contexts.
\newblock {\em IEEE transactions on pattern analysis and machine intelligence},
  24(4):509--522, 2002.

\bibitem{bookstein1989principal}
Fred~L. Bookstein.
\newblock Principal warps: Thin-plate splines and the decomposition of
  deformations.
\newblock {\em IEEE Transactions on pattern analysis and machine intelligence},
  11(6):567--585, 1989.

\bibitem{brouet2012design}
Remi Brouet, Alla Sheffer, Laurence Boissieux, and Marie-Paule Cani.
\newblock Design preserving garment transfer.
\newblock {\em ACM Transactions on Graphics}, 31(4):Article--No, 2012.

\bibitem{chen2016synthesizing}
Wenzheng Chen, Huan Wang, Yangyan Li, Hao Su, Zhenhua Wang, Changhe Tu, Dani
  Lischinski, Daniel Cohen-Or, and Baoquan Chen.
\newblock Synthesizing training images for boosting human 3d pose estimation.
\newblock In {\em International Conference on 3D Vision}, pages 479--488. IEEE,
  2016.

\bibitem{choi2021viton}
Seunghwan Choi, Sunghyun Park, Minsoo Lee, and Jaegul Choo.
\newblock Viton-hd: High-resolution virtual try-on via misalignment-aware
  normalization.
\newblock In {\em Proceedings of the IEEE/CVF conference on computer vision and
  pattern recognition}, pages 14131--14140, 2021.

\bibitem{choi2018stargan}
Yunjey Choi, Minje Choi, Munyoung Kim, Jung-Woo Ha, Sunghun Kim, and Jaegul
  Choo.
\newblock Stargan: Unified generative adversarial networks for multi-domain
  image-to-image translation.
\newblock In {\em Conference on Computer Vision and Pattern Recognition}, pages
  8789--8797, 2018.

\bibitem{chopra2021zflow}
Ayush Chopra, Rishabh Jain, Mayur Hemani, and Balaji Krishnamurthy.
\newblock Zflow: Gated appearance flow-based virtual try-on with 3d priors.
\newblock In {\em Proceedings of the IEEE/CVF International Conference on
  Computer Vision}, pages 5433--5442, 2021.

\bibitem{cui2021dressing}
Aiyu Cui, Daniel McKee, and Svetlana Lazebnik.
\newblock Dressing in order: Recurrent person image generation for pose
  transfer, virtual try-on and outfit editing.
\newblock In {\em Proceedings of the IEEE/CVF International Conference on
  Computer Vision}, pages 14638--14647, 2021.

\bibitem{devlin2018bert}
Jacob Devlin, Ming-Wei Chang, Kenton Lee, and Kristina Toutanova.
\newblock Bert: Pre-training of deep bidirectional transformers for language
  understanding.
\newblock In {\em Annual Conference of the North American Chapter of the
  Association for Computational Linguistics (NAACL)}, 2018.

\bibitem{dong2019fw}
Haoye Dong, Xiaodan Liang, Xiaohui Shen, Bowen Wu, Bing-Cheng Chen, and Jian
  Yin.
\newblock Fw-gan: Flow-navigated warping gan for video virtual try-on.
\newblock In {\em International Conference on Computer Vision}, pages
  1161--1170, 2019.

\bibitem{dosovitskiy2020image}
Alexey Dosovitskiy, Lucas Beyer, Alexander Kolesnikov, Dirk Weissenborn,
  Xiaohua Zhai, Thomas Unterthiner, Mostafa Dehghani, Matthias Minderer, Georg
  Heigold, Sylvain Gelly, et~al.
\newblock An image is worth 16x16 words: Transformers for image recognition at
  scale.
\newblock In {\em International Conference on Learning Representations}, 2021.

\bibitem{ehara2006texture}
Jun Ehara and Hideo Saito.
\newblock Texture overlay for virtual clothing based on pca of silhouettes.
\newblock In {\em IEEE/ACM International Symposium on Mixed and Augmented
  Reality}, pages 139--142. Citeseer, 2006.

\bibitem{fele2022c}
Benjamin Fele, Ajda Lampe, Peter Peer, and Vitomir Struc.
\newblock C-vton: Context-driven image-based virtual try-on network.
\newblock In {\em Proceedings of the IEEE/CVF Winter Conference on Applications
  of Computer Vision}, pages 3144--3153, 2022.

\bibitem{fincato2022transform}
Matteo Fincato, Marcella Cornia, Federico Landi, Fabio Cesari, and Rita
  Cucchiara.
\newblock Transform, warp, and dress: a new transformation-guided model for
  virtual try-on.
\newblock {\em ACM Transactions on Multimedia Computing, Communications, and
  Applications (TOMM)}, 18(2):1--24, 2022.

\bibitem{fincato2021viton}
Matteo Fincato, Federico Landi, Marcella Cornia, Fabio Cesari, and Rita
  Cucchiara.
\newblock Viton-gt: an image-based virtual try-on model with geometric
  transformations.
\newblock In {\em 2020 25th International Conference on Pattern Recognition
  (ICPR)}, pages 7669--7676. IEEE, 2021.

\bibitem{ge2021disentangled}
Chongjian Ge, Yibing Song, Yuying Ge, Han Yang, Wei Liu, and Ping Luo.
\newblock Disentangled cycle consistency for highly-realistic virtual try-on.
\newblock In {\em Proceedings of the IEEE/CVF conference on computer vision and
  pattern recognition}, pages 16928--16937, 2021.

\bibitem{gu2020toward}
Xiaoling Gu, Jun Yu, Yongkang Wong, and Mohan~S Kankanhalli.
\newblock Toward multi-modal conditioned fashion image translation.
\newblock {\em IEEE Transactions on Multimedia}, 2020.

\bibitem{guan2012drape}
Peng Guan, Loretta Reiss, David~A Hirshberg, Alexander Weiss, and Michael~J
  Black.
\newblock Drape: Dressing any person.
\newblock {\em ACM Transactions on Graphics}, 31(4):1--10, 2012.

\bibitem{gundogdu2019garnet}
Erhan Gundogdu, Victor Constantin, Amrollah Seifoddini, Minh Dang, Mathieu
  Salzmann, and Pascal Fua.
\newblock Garnet: A two-stream network for fast and accurate 3d cloth draping.
\newblock In {\em International Conference on Computer Vision}, pages
  8739--8748, 2019.

\bibitem{han2018viton}
Xintong Han, Zuxuan Wu, Zhe Wu, Ruichi Yu, and Larry~S Davis.
\newblock Viton: An image-based virtual try-on network.
\newblock In {\em Conference on Computer Vision and Pattern Recognition}, pages
  7543--7552, 2018.

\bibitem{he2022style}
Sen He, Yi-Zhe Song, and Tao Xiang.
\newblock Style-based global appearance flow for virtual try-on.
\newblock In {\em Proceedings of the IEEE/CVF Conference on Computer Vision and
  Pattern Recognition}, pages 3470--3479, 2022.

\bibitem{issenhuth2020not}
Thibaut Issenhuth, J{\'e}r{\'e}mie Mary, and Cl{\'e}ment Calauzenes.
\newblock Do not mask what you do not need to mask: a parser-free virtual
  try-on.
\newblock In {\em Computer Vision--ECCV 2020: 16th European Conference,
  Glasgow, 23--28, 2020}, pages 619--635. Springer, 2020.

\bibitem{jaccard1912distribution}
Paul Jaccard.
\newblock The distribution of the flora in the alpine zone. 1.
\newblock {\em New phytologist}, 11(2):37--50, 1912.

\bibitem{jetchev2017conditional}
Nikolay Jetchev and Urs Bergmann.
\newblock The conditional analogy gan: Swapping fashion articles on people
  images.
\newblock In {\em International Conference on Computer Vision Workshops}, pages
  2287--2292, 2017.

\bibitem{johnson2016perceptual}
Justin Johnson, Alexandre Alahi, and Li Fei-Fei.
\newblock Perceptual losses for real-time style transfer and super-resolution.
\newblock In {\em European conference on computer vision}, pages 694--711.
  Springer, 2016.

\bibitem{khan2022transformers}
Salman Khan, Muzammal Naseer, Munawar Hayat, Syed~Waqas Zamir, Fahad~Shahbaz
  Khan, and Mubarak Shah.
\newblock Transformers in vision: A survey.
\newblock {\em ACM computing surveys (CSUR)}, 54(10s):1--41, 2022.

\bibitem{lahner2018deepwrinkles}
Zorah Lahner, Daniel Cremers, and Tony Tung.
\newblock Deepwrinkles: Accurate and realistic clothing modeling.
\newblock In {\em European Conference on Computer Vision}, pages 667--684,
  2018.

\bibitem{lee2022high}
Sangyun Lee, Gyojung Gu, Sunghyun Park, Seunghwan Choi, and Jaegul Choo.
\newblock High-resolution virtual try-on with misalignment and
  occlusion-handled conditions.
\newblock In {\em European Conference on Computer Vision}, pages 204--219.
  Springer, 2022.

\bibitem{li2021toward}
Kedan Li, Min~Jin Chong, Jeffrey Zhang, and Jingen Liu.
\newblock Toward accurate and realistic outfits visualization with attention to
  details.
\newblock In {\em Proceedings of the IEEE/CVF conference on computer vision and
  pattern recognition}, pages 15546--15555, 2021.

\bibitem{liu2019swapgan}
Yu Liu, Wei Chen, Li Liu, and Michael~S Lew.
\newblock Swapgan: A multistage generative approach for person-to-person
  fashion style transfer.
\newblock {\em IEEE Transactions on Multimedia}, 21(9):2209--2222, 2019.

\bibitem{liu2022breaking}
Yahui Liu, Bin Ren, Yue Song, Wei Bi, Nicu Sebe, and Wei Wang.
\newblock Breaking the chain of gradient leakage in vision transformers.
\newblock {\em arXiv preprint arXiv:2205.12551}, 2022.

\bibitem{lowe2004distinctive}
David~G Lowe.
\newblock Distinctive image features from scale-invariant keypoints.
\newblock {\em International journal of computer vision}, 60(2):91--110, 2004.

\bibitem{men2020controllable}
Yifang Men, Yiming Mao, Yuning Jiang, Wei-Ying Ma, and Zhouhui Lian.
\newblock Controllable person image synthesis with attribute-decomposed gan.
\newblock In {\em Conference on Computer Vision and Pattern Recognition}, pages
  5084--5093, 2020.

\bibitem{mikolajczyk2002affine}
Krystian Mikolajczyk and Cordelia Schmid.
\newblock An affine invariant interest point detector.
\newblock In {\em European Conference on Computer Vision}, pages 128--142,
  2002.

\bibitem{minar2020cp}
MR Minar, TT Tuan, H Ahn, P Rosin, and YK Lai.
\newblock Cp-vton+: Clothing shape and texture preserving image-based virtual
  try-on.
\newblock In {\em Conference on Computer Vision and Pattern Recognition
  Workshops}, volume~2, page~11, 2020.

\bibitem{morelli2022dress}
Davide Morelli, Matteo Fincato, Marcella Cornia, Federico Landi, Fabio Cesari,
  and Rita Cucchiara.
\newblock Dress code: High-resolution multi-category virtual try-on.
\newblock In {\em Computer Vision--ECCV 2022: 17th European Conference, Tel
  Aviv, Israel, October 23--27, 2022, Proceedings, Part VIII}, pages 345--362,
  2022.

\bibitem{ott2019fairseq}
Myle Ott, Sergey Edunov, Alexei Baevski, Angela Fan, Sam Gross, Nathan Ng,
  David Grangier, and Michael Auli.
\newblock fairseq: A fast, extensible toolkit for sequence modeling.
\newblock In {\em Conference of the North American Chapter of the Association
  for Computational Linguistics :Demonstrations}, pages 48--53, 2019.

\bibitem{pons2017clothcap}
Gerard Pons-Moll, Sergi Pujades, Sonny Hu, and Michael~J Black.
\newblock Clothcap: Seamless 4d clothing capture and retargeting.
\newblock {\em ACM Transactions on Graphics}, 36(4):1--15, 2017.

\bibitem{ren2023masked}
Bin Ren, Yahui Liu, Yue Song, Wei Bi, Rita Cucchiara, Nicu Sebe, and Wei Wang.
\newblock Masked jigsaw puzzle: A versatile position embedding for vision
  transformers.
\newblock In {\em Proceedings of the IEEE/CVF Conference on Computer Vision and
  Pattern Recognition}, pages 20382--20391, 2023.

\bibitem{ren2021cascaded}
Bin Ren, Hao Tang, and Nicu Sebe.
\newblock Cascaded cross mlp-mixer gans for cross-view image translation.
\newblock {\em British Machine Vision Conference}, 2021.

\bibitem{rocco2017convolutional}
Ignacio Rocco, Relja Arandjelovic, and Josef Sivic.
\newblock Convolutional neural network architecture for geometric matching.
\newblock In {\em Computer Vision and Pattern Recognition}, pages 6148--6157,
  2017.

\bibitem{salimans2016improved}
Tim Salimans, Ian~J Goodfellow, Wojciech Zaremba, Vicki Cheung, Alec Radford,
  and Xi Chen.
\newblock Improved techniques for training gans.
\newblock In {\em Neural Information Processing Systems}, 2016.

\bibitem{schlemper2019attention}
Jo Schlemper, Ozan Oktay, Michiel Schaap, Mattias Heinrich, Bernhard Kainz, Ben
  Glocker, and Daniel Rueckert.
\newblock Attention gated networks: Learning to leverage salient regions in
  medical images.
\newblock {\em Medical image analysis}, 53:197--207, 2019.

\bibitem{schmid1997local}
Cordelia Schmid and Roger Mohr.
\newblock Local grayvalue invariants for image retrieval.
\newblock {\em IEEE transactions on pattern analysis and machine intelligence},
  19(5):530--535, 1997.

\bibitem{sekhavat2016privacy}
Yoones~A Sekhavat.
\newblock Privacy preserving cloth try-on using mobile augmented reality.
\newblock {\em IEEE Transactions on Multimedia}, 19(5):1041--1049, 2016.

\bibitem{sekine2014virtual}
Masahiro Sekine, Kaoru Sugita, Frank Perbet, Bj{\"o}rn Stenger, and Masashi
  Nishiyama.
\newblock Virtual fitting by single-shot body shape estimation.
\newblock In {\em Int. Conf. on 3D Body Scanning Technologies}, pages 406--413.
  Citeseer, 2014.

\bibitem{strubell2018linguistically}
Emma Strubell, Patrick Verga, Daniel Andor, David Weiss, and Andrew McCallum.
\newblock Linguistically-informed self-attention for semantic role labeling.
\newblock In {\em Conference on Empirical Methods in Natural Language
  Processing}, pages 5027--5038, 2018.

\bibitem{tang2020bipartite}
Hao Tang, Song Bai, Philip~HS Torr, and Nicu Sebe.
\newblock Bipartite graph reasoning gans for person image generation.
\newblock In {\em British Machine Vision Conference}, 2020.

\bibitem{tang2020xinggan}
Hao Tang, Song Bai, Li Zhang, Philip~HS Torr, and Nicu Sebe.
\newblock Xinggan for person image generation.
\newblock In {\em European Conference on Computer Vision}, pages 717--734.
  Springer, 2020.

\bibitem{tolstikhin2021mlp}
Ilya~O Tolstikhin, Neil Houlsby, Alexander Kolesnikov, Lucas Beyer, Xiaohua
  Zhai, Thomas Unterthiner, Jessica Yung, Andreas Steiner, Daniel Keysers,
  Jakob Uszkoreit, et~al.
\newblock Mlp-mixer: An all-mlp architecture for vision.
\newblock {\em NeurIPS}, 34, 2021.

\bibitem{tsai2019MULT}
Yao-Hung~Hubert Tsai, Shaojie Bai, Paul~Pu Liang, J.~Zico Kolter,
  Louis-Philippe Morency, and Ruslan Salakhutdinov.
\newblock Multimodal transformer for unaligned multimodal language sequences.
\newblock In {\em Annual Meeting of the Association for Computational
  Linguistics}, 7 2019.

\bibitem{vaswani2017attention}
Ashish Vaswani, Noam Shazeer, Niki Parmar, Jakob Uszkoreit, Llion Jones,
  Aidan~N Gomez, Lukasz Kaiser, and Illia Polosukhin.
\newblock Attention is all you need.
\newblock In {\em Neural Information Processing Systems}, 2017.

\bibitem{wang2018toward}
Bochao Wang, Huabin Zheng, Xiaodan Liang, Yimin Chen, Liang Lin, and Meng Yang.
\newblock Toward characteristic-preserving image-based virtual try-on network.
\newblock In {\em European Conference on Computer Vision}, pages 589--604,
  2018.

\bibitem{wang2018non}
Xiaolong Wang, Ross Girshick, Abhinav Gupta, and Kaiming He.
\newblock Non-local neural networks.
\newblock In {\em Conference on Computer Vision and Pattern Recognition}, pages
  7794--7803, 2018.

\bibitem{wang2004image}
Zhou Wang, Alan~C Bovik, Hamid~R Sheikh, and Eero~P Simoncelli.
\newblock Image quality assessment: from error visibility to structural
  similarity.
\newblock {\em IEEE transactions on image processing}, 13(4):600--612, 2004.

\bibitem{xu2021virtual}
Jun Xu, Yuanyuan Pu, Rencan Nie, Dan Xu, Zhengpeng Zhao, and Wenhua Qian.
\newblock Virtual try-on network with attribute transformation and local
  rendering.
\newblock {\em IEEE Transactions on Multimedia}, 2021.

\bibitem{yang2020towards}
Han Yang, Ruimao Zhang, Xiaobao Guo, Wei Liu, Wangmeng Zuo, and Ping Luo.
\newblock Towards photo-realistic virtual try-on by adaptively
  generating-preserving image content.
\newblock In {\em Conference on Computer Vision and Pattern Recognition}, pages
  7850--7859, 2020.

\bibitem{yu2019vtnfp}
Ruiyun Yu, Xiaoqi Wang, and Xiaohui Xie.
\newblock Vtnfp: An image-based virtual try-on network with body and clothing
  feature preservation.
\newblock In {\em Proceedings of the IEEE/CVF International Conference on
  Computer Vision}, pages 10511--10520, 2019.

\bibitem{yuan2013mixed}
Miaolong Yuan, Ishtiaq~Rasool Khan, Farzam Farbiz, Susu Yao, Arthur Niswar, and
  Min-Hui Foo.
\newblock A mixed reality virtual clothes try-on system.
\newblock {\em IEEE Transactions on Multimedia}, 15(8):1958--1968, 2013.

\bibitem{zhang2018unreasonable}
Richard Zhang, Phillip Isola, Alexei~A Efros, Eli Shechtman, and Oliver Wang.
\newblock The unreasonable effectiveness of deep features as a perceptual
  metric.
\newblock In {\em Conference on Computer Vision and Pattern Recognition}, pages
  586--595, 2018.

\bibitem{zhu2017your}
Shizhan Zhu, Raquel Urtasun, Sanja Fidler, Dahua Lin, and Chen Change~Loy.
\newblock Be your own prada: Fashion synthesis with structural coherence.
\newblock In {\em International Conference on Computer Vision}, pages
  1680--1688, 2017.

\end{thebibliography}
}

\end{document}


\title{Correspondence Transformation and Difference Alignment GANs for Person Image Generation\\
\vspace{5pt}
--Supplementary Document--}

\author{First Author\\
Institution1\\
Institution1 address\\
{\tt\small firstauthor@i1.org}
\and
Second Author\\
Institution2\\
First line of institution2 address\\
{\tt\small secondauthor@i2.org}
}

\maketitle

\input{6supple}

\clearpage
{\small
\bibliographystyle{ieee_fullname}
\bibliography{egbib}
}

\clearpage